\documentclass[a4paper]{ETHpaper}

\title{Predicting Sequences of Traversed Nodes in Graphs using Network Models with Multiple Higher Orders}

\author{Christoph Gote\textsuperscript{\rm 1}, Giona Casiraghi\textsuperscript{\rm 1}, Frank Schweitzer\textsuperscript{\rm 1}, and Ingo Scholtes\textsuperscript{\rm 2,3}}
\authoralternative{Christoph Gote, Giona Casiraghi, Frank Schweitzer, and Ingo Scholtes}
\address{\textsuperscript{1}Chair of Systems Design, ETH Zurich, Zurich, Switzerland\\
	\textsuperscript{2}Chair of Data Analytics, University of Wuppertal, Wuppertal, Germany\\
	\textsuperscript{3}Data Analytics Group, University of Zurich, Zurich, Switzerland}

\usepackage{amsmath}
\usepackage{blkarray}
\usepackage{tabularx}
\usepackage{makecell}
\usepackage{cleveref}
\usepackage{booktabs}
\usepackage[english]{babel}

\usepackage{tikz}
\usetikzlibrary{positioning, calc}
\usetikzlibrary{decorations}
\usetikzlibrary{arrows.meta}
\usetikzlibrary{arrows}
\usetikzlibrary{decorations.shapes}
\usepackage{pgfplots}
\usepgfplotslibrary{groupplots}
\usepackage{adjustbox}
\usepackage[colorinlistoftodos,disable]{todonotes}
\newcolumntype{C}{>{\centering\arraybackslash}X}
\newcolumntype{R}{>{\raggedleft\arraybackslash}X}
\usepackage{fontawesome}
\usepackage{subcaption}
\DeclareMathAlphabet\mathbfcal{OMS}{cmsy}{b}{n}
\usepackage{nicefrac}
\pgfplotsset{compat=newest}
\usepackage[square, sort&compress, numbers]{natbib}
\definecolor{colorSG}{RGB}{168,50,45}
\definecolor{customY}{HTML}{FBB13C}
\definecolor{customG}{HTML}{218380}
\definecolor{customT}{HTML}{73D2DE}
\definecolor{customW}{HTML}{FCFCFF}
\definecolor{customB}{HTML}{2E5EAA}
\definecolor{customP}{HTML}{5B4E77}
\definecolor{gitRedFaded}{HTML}{FFEEF0}
\definecolor{gitGreenFaded}{HTML}{E6FFED}
\definecolor{gitRed}{HTML}{FFDCE0}
\definecolor{gitGreen}{HTML}{CDFFD8}
\definecolor{gitRedFull}{HTML}{CB2431}
\definecolor{gitGreenFull}{HTML}{2CBE4E}

\newcommand\mydots{\makebox[.6em][c]{\hspace{-.4mm}.\hfil.\hfil.\hspace{.4mm}}}

\newcommand{\prc}[1]{Q^{#1}_{\phantom{r}}}
\newcommand{\parm}[2]{n_{#1}^{#2}}
\newcommand{\dat}[2]{\hat n_{#1}^{#2}}

\DeclareMathOperator*{\argmin}{arg\,min}

\def\methodname/{MOGen}

\begin{document}

\maketitle

\begin{abstract}
	We propose a novel sequence prediction method for sequential data capturing node traversals in graphs.
	Our method builds on a statistical modelling framework that combines multiple higher-order network models into a single multi-order model.
	We develop a technique to fit such multi-order models in empirical sequential data and to select the optimal maximum order.
	Our framework facilitates both next-element and full sequence prediction given a sequence-prefix of any length.
	We evaluate our model based on six empirical data sets containing sequences from website navigation as well as public transport systems.
	The results show that our method out-performs state-of-the-art algorithms for next-element prediction.
	We further demonstrate the accuracy of our method during out-of-sample sequence prediction and validate that our method can scale to data sets with millions of sequences. 
\end{abstract}

\begin{figure*}[!htb]
	\centering
	\scalebox{0.9}{
	\input{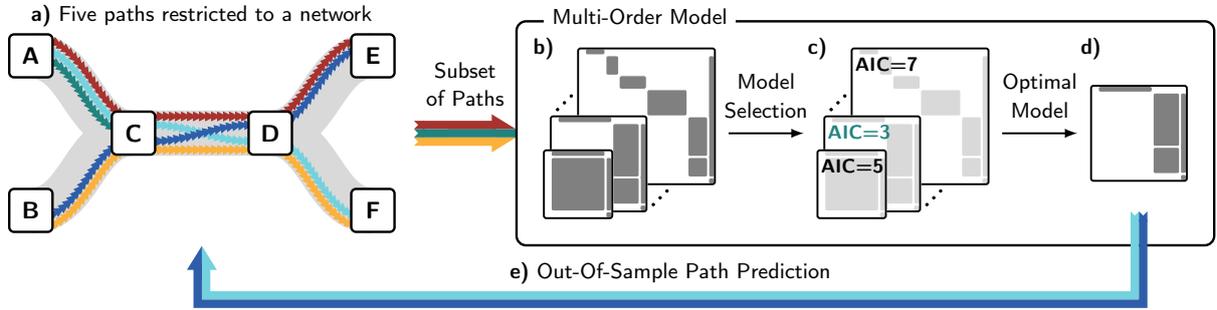}
	}
	\caption{Overview of the proposed framework to predict paths in networks: Using paths observed in a network (a), \methodname/ learns a multi-order generative model (b) that accounts for (i) sequential patterns in the sequence of traversed nodes, (ii) endpoints of paths, and (iii) length distributions of paths in a training set. An information-based model selection algorithm (c) yields an optimal generative model (d) that can be used for an out-of-sample prediction of paths in a validation set (e).}
	\label{fig:infographic}
	\end{figure*}

\noindent The analysis of data on complex networks provides essential insights into the structure and dynamics of social, technical, and biological systems.
Apart from data that capture the \emph{topology} of networked systems, i.e., which elements are directly connected via links, we increasingly have access to time-resolved, sequential data on \emph{paths or trajectories in networks}.
Examples include clickstream data generated by users who follow hyperlinks in information networks, data on information cascades propagating along friendship relations in online social networks, or temporal data capturing the travel routes of passengers in a transportation network.
We can represent such path data as a multi set $S$ of node sequences $v_1 \rightarrow v_2 \rightarrow \dots \rightarrow v_l$ of variable-length $l$, in a graph $G=(V,E)$ under the constraint that nodes $v_i \in V$ and transitions $(v_{i}, v_{i+1}) \in E$ for $i=1, \ldots, l-1$. 
That is, different from general categorical sequence data, paths are constrained sequences of nodes where consecutive nodes are connected by a link in an underlying graph (cf. \cref{fig:infographic}a).
Such data provide two complementary dimensions of information on networked systems: (i) the network, which determines, e.g., which connections passengers can use or which hyperlinks users can click, and (ii) path data, which determine along which specific paths passengers actually travel or how users sequentially navigate through web pages.

Recent works have shown that data on paths in networked systems expose \emph{higher-order dependencies} between nodes, i.e., correlations in the sequence of nodes traversed on paths that cannot be explained by the network topology \cite{Lambiotte2019_Understanding,Torres2020_NetworkRepresentations,Battiston2020_Networks}.
The modelling of patterns in such data is both a topical and a challenging problem.
On the one hand, data on paths in networks rule out state-of-the-art sequential pattern mining algorithms that do not account for (i) the constraint that an underlying network limits the sequences of traversed nodes, and (ii) the fact that path data typically consists of many short node sequences with varying lengths.
On the other hand, path data invalidates network analysis and graph mining techniques that neglect sequential patterns in the sequence of traversed nodes.

Addressing this challenge, recent works in data science and network analysis have proposed to use \emph{higher-order models for paths in networks}, which generalise graph representations to higher-dimensional models that capture sequential patterns in paths.
Such models capture the \emph{topology of paths} rather than the \emph{topology of links}, e.g., by representing a path $v_1 \rightarrow v_2 \rightarrow v_3$ as a link between two \emph {second-order nodes} $(v_1, v_2)$ and $(v_2, v_3)$ in a second-order network model. 
This representation generalises the concept of a link in standard graph representations, which can be viewed as a path of length two, to $k$-th order networks where links represent paths of length $k+1$.
Such generalisations of graph models to higher orders improve our ability to cluster and rank nodes~\cite{Rosvall2014_Memory,Xu2016,scholtes2017network}, detect anomalies~\cite{LaRock2019_HYPA}, and open new perspectives for network embedding~\cite{Saebi2019_HONEM,Belth2019_WhenToRemember}.
However, the modelling assumptions underlying those works hinder an application to the supervised prediction of paths, which is the motivation for our work.

\textbf{Contributions:} To address this issue, we introduce \methodname/, a method to learn \emph{generative multi-order models} in large collections of variable-length paths in networks.
Our contributions are as follows.
(i) Extending recent research on higher-, variable-, and multi-order models of paths, our framework combines Markov chains of multiple orders into a single generative model that accounts for (a) patterns in the sequence of traversed nodes, and (b) the constraint that sequences of traversed nodes are limited to an underlying network.
(ii) Exposing a---to the best of our knowledge---previously unknown relation between higher-order network models and matrix representations of multi-layer networks~\cite{Kivela_MultiLayer}, we show how our models can be formalised in terms of block adjacency and transition matrices.
(iii) We propose an information-based model selection algorithm that detects the correct maximum model order needed to model sequential patterns in a given set of paths.
We evaluate this algorithm in empirical data and show that it yields generalisable models that neither under- nor overfit sequential patterns.
(iv) We develop a supervised learning technique to predict variable-length paths in networks. 
Different from existing algorithms, our method supports both predicting the next node based on a prefix of varying length (next-element prediction) as well as the out-of-sample prediction of full paths based on a training set.
We apply our method to six data sets on human clickstreams and passenger itineraries in transportation networks and demonstrate its superior performance through a comparison to state-of-the-art sequence prediction algorithms.
An overview of our method is shown in \cref{fig:infographic}.

The remainder of this article is structured as follows.
First, we summarise the state-of-the-art the modelling of categorical sequences and paths and identify the research gap that is addressed by our work.
Subsequently we introduce our method.
We then experimentally validate our method in six empirical data sets.
We apply our method to the supervised prediction of paths, compare its performance to state-of-the-art sequence prediction algorithms, and evaluate our model selection algorithm.
Finally, we summarise our conclusions.

\section{Related Work}
\label{sec:relatedwork}

The modelling of categorical sequences has been addressed in the data mining and machine learning community.
While we refer to \cite{next_element_review,Fournier2017_Survey} for an exhaustive overview on sequential pattern mining, here we limit our discussion of related work to (i) methods to model and predict categorical sequences, and (ii) works that explicitly address the modelling of paths in a network.

\paragraph{Sequence Modelling and Prediction}
The modelling and prediction of categorical sequence data have important applications, e.g., in text mining, machine translation, DNA sequence analysis, clickstream analysis and human behavioural modelling.
A large body of works has applied probabilistic graphical models like, e.g., Petri nets, hidden Markov models, decision trees, as well as higher- or variable-order Markov chains~\cite{Chierichetti2012_Markovian,West2012,Singer2014_Memory,Benson2017_SpaceyRW} to model patterns in categorical sequences~\cite{next_element_review,Fournier2017_Survey}.
Such sequence modelling techniques facilitate \emph{next-element prediction}, which seeks to predict the next element(s) in a given sequence based on a model learned in training sequences.
This problem occurs in a number of applications, e.g., recommender systems~\cite{frias2002prediction}, speculative Web caching~\cite{bestavros1995using}, or purchase prediction in eCommerce~\cite{Kraus2019_Purchase}.
Such applications have been addressed with (higher-order) Markov chains that predict the next element in a sequence based on the previous $k$ elements~\cite{deshpande2004selective,gunduz2003web,bernhard2016clickstream,AKOM,PPM}.
Other methods use Petri nets \cite{FHM,SM}, decision trees \cite{CPT+,ETMd,Indulpet}, graph and network models \cite{DG,TDAG}, spectral learning \cite{SL}, as well as neural networks and automata \cite{next_element_review}.
An advantage of those methods is that they can be applied to general data on categorical sequences.
However, data on paths in networks have characteristics that violate the assumptions of some of those techniques. 
First, paths are \emph{constrained} categorical sequences insofar as an underlying network restricts the possible sequences of traversed nodes.
In terms of sequential pattern mining, this has the consequence that some frequently traversed node sequences are trivially due to the topology of the underlying network, while others can not be explained by the network alone.
Second, different from data that contain a small number of long categorical sequences, path data typically confront us with a large unordered collection of \emph{short and variable-length sequences}, where each sequence is one independent observation of, e.g., a user clickstream in the Web or an itinerary of a passenger in a transportation network.
The structure of such data complicates the application of higher-order Markov chains with large orders $k$, which treat the first $k$ nodes in each path as a prefix for the following transitions.

\paragraph{Higher-Order Models of Paths in Networks}
Several recent works highlight the need for \emph{higher-order network models} to account for the special characteristics of sequential data on paths.
\cite{Lambiotte2019_Understanding} summarise the challenges of this research at the intersection of network analysis, data mining, and machine learning.
The key idea behind those works is the use of higher-dimensional graph representations.
The resulting higher-order network models resemble the state space extension that is the basis of higher-order Markov chains or high-dimensional De Bruijn graphs~\cite{bruijngraph}.
Building on this idea, \cite{Rosvall2014_Memory} use a generalisation of the flow compression algorithm InfoMap \cite{Rosvall2008_Maps} to higher-order network models to detect temporal cluster structures in sequential data on networks.
\cite{Scholtes2014_CausalityDriven} show that the spectral properties of higher-order graph Laplacians can be used to improve the prediction of diffusion processes based on time-stamped and sequential data on networks.
A number of works have generalised node centrality measures and ranking algorithms, demonstrating that higher-order models of paths improve our ability to identify important nodes in networks \cite{Rosvall2014_Memory,Xu2016,scholtes2017network,Scholtes2016_HigherOrder}.
Recent works used such models to capture event sequences in time-stamped network data~\cite{Mellor2019_EventGraphs}, detect anomalies in data on paths in networks \cite{LaRock2019_HYPA}, or propose new network embedding and representation learning techniques~\cite{Saebi2019_HONEM,Belth2019_WhenToRemember}.

Those works generated insights into the topology of networked systems that cannot be obtained with standard network analysis and graph mining techniques. 
However, the modelling assumptions underlying those works also limit their application in many important settings.
First, existing works have focused on higher-order models for patterns in the sequence of traversed nodes, rather than \emph{generative models} that additionally model the endpoints of paths \cite{Lambiotte2019_Understanding}.
This limits the ability to predict variable-length paths, an important supervised learning problem that is the key motivation for our work.
A second limitation is the choice of model selection techniques to learn \emph{generalisable higher-order models} that provide an optimal balance between model complexity and explanatory power.
Existing works either used heuristic techniques \cite{Xu2016,Rosvall2014_Memory} or analytical methods that rely on additional model characteristics like, e.g., a nested structure \cite{scholtes2017network} that do not apply to generative models.
Moreover, most existing works used models with a single order $k$, which limits their application in paths that exhibit correlations at multiple scales at once.
While both model selection and sequence prediction have been addressed in the modelling of categorical sequences, we note that---due to the special characteristics outlined above---those methods cannot be immediately applied to path data.

In the remainder of this article, we develop a generative multi-order modelling framework for paths in networks that closes this gap.
Different from state-of-the-art sequence modelling and prediction algorithms, our method explicitly addresses the modelling of unordered collections of variable-length node sequences constrained by a network.
Different from previous works on higher- and multi-order models \cite{Rosvall2014_Memory,Xu2016,scholtes2017network}, we focus on a generative model that accounts for the endpoints of paths and is thus suitable for prediction tasks.
By combining models of multiple higher-order into a multi-layered model structure, our method captures sequential patterns at multiple length scales simultaneously.
We further develop an information-based model selection algorithm that allows detecting the optimal maximum order of the model without requiring the nestedness property that is the basis for the method proposed in \cite{scholtes2017network}.

\section{Multi-Order Generative Model for Paths}
\label{sec:method}

We first formally define path data and present the generative modelling framework and model selection algorithm that are the main contributions of our work.
We assume that we are given a multi-set of $m$ paths $S = \{t_1, \dots, t_m\}$ on a network $G = (V, E)$ with nodes $V$ and (directed) links $E \subseteq V \times V$.
Each path $t_i = v_1 \rightarrow v_2 \rightarrow \dots \rightarrow v_{l_i}$ is an ordered sequence of traversed nodes under the constraint that $(v_{j}, v_{j+1}) \in E$ for $j \in [1, l_i-1]$.
We assume that paths in $S$ have varying lengths $l_i$ and we denote the maximum path length as $l_{max}$.
Using a simple network-based model for the sequences of nodes traversed by paths, we could assume that paths are generated by a memoryless random walk that corresponds to a first-order Markov chain.
The nodes $v \in V$ of network $G$ correspond to the state space of such a Markov chain, while links $(v,w) \in E$ in the network are associated with transition probabilities.
This assumes that the next node $v_{i+1}$ traversed by a path only depends on (i) the currently visited node $v_{i}$, and (ii) the links that determine which state transitions are possible.
To capture higher-order dependencies in the sequence of traversed nodes that are not due to the network topology, we can generalise this model to a $k$-th order Markov chain, where the next node $v_{i+i}$ depends on the $k$ previously visited nodes $v_{i-k}, \mydots, v_{i}$.
We can view this model as a memoryless random walk in a $k$-th order network $G^{k}=(V^{k}, E^{k})$, where (i) each $k$-th order node $(v_1, \mydots, v_k) \in V^k$ is a sequence of $k$ nodes where consecutive nodes are connected by links, and (ii) each $k$-th order link connects two higher-order nodes $(v_1, \mydots, v_k)$ and $(w_1, \mydots, w_k)$ such that $w_{i}=v_{i-1} \in V$.
We note that a $k$-th order network can be constructed as the line graph of a $k-1$-th order network, a construction that adapts the construction of De Bruijn graphs to models for paths in a network~\cite{bruijngraph,scholtes2017network}.
Using this modelling approach, a path $v_1 \rightarrow v_2 \rightarrow \dots \rightarrow v_{l}$ corresponds to $l-k$ transitions along the edges of a $k$-th order network, which gives rise to the following sequence of $l-k+1$ higher-order nodes:
\begin{equation*}
	(v_1, \mydots, v_{k}) \rightarrow (v_2, \mydots, v_{k+1}) \rightarrow \dots \rightarrow (v_{l-k+1}, \mydots, v_{l})
\end{equation*}

\subsection{Multi-Order Models of Paths}
\label{sec:method:multiorder}

Referring to the special characteristics of variable-length paths in $S$, the modelling approach above has a number of limitations.
(i) Using a $k$-th order model, we can neither use the first $k$ nodes in each path to fit the transition probabilities of the model, nor can we use the model to perform a next-element prediction for the first $k$ nodes on a path.
(ii) The fact that the number of observations used to fit the transition probabilities of a $k$-th order model differs across different orders $k$ complicates model selection, i.e., the selection of the optimal order $k$ to model a data set. 
(iii) The model neglects information on the start and end point of nodes, which hinders its application to an out-of-sample prediction of full paths, i.e. from start to end.
(iv) Finally, a model with a single order $k$ is limited to capture sequential patterns of a single length $k$, which is why recent works argued for variable- and multi-order modelling techniques~\cite{scholtes2017network,Xu2016,AKOM}.

To overcome those limitations we introduce a generative model that combines (i) transitions between higher-order models of multiple orders up to a maximum order $K$ and (ii) special transitions that capture the start and end points of paths.
For this, we introduce a special \emph{initial state} $*$, with transitions $* \rightarrow v$ to first-order nodes $v \in V$ that mark the start of a path in node $v$.
For all orders $1 \leq k \leq K$, we further add transitions from $k$-th order nodes $(v_1, \mydots, v_k) \rightarrow \dagger$ to a special \emph{terminal state} $\dagger$, which captures the termination of a path.
With this extension, a path $v_1 \rightarrow v_2 \rightarrow \dots \rightarrow v_{l}$ corresponds to $l+2$ transitions in a \emph{multi-order network} with maximum order $K$, which gives rise to the following sequence of $l$ higher-order nodes
\begin{align*}
	* \rightarrow v_1 \rightarrow (v_1, v_2) \rightarrow (v_1, v_2, v_3) \rightarrow \dots\rightarrow (v_{l-K+1}, \mydots, v_{l}) \rightarrow \dagger
	%\label{eq:path_representation}
\end{align*}
with two additional transitions from/to the special states $*$ and $\dagger$ to model the start and end of the path.
Considering the coding of memory prefixes in the extended state space of higher-order Markov chains, we note that each transition in this multi-order model with maximum order $K$ increases the memory prefix by one node, up to a maximum memory of length $K$.
In addition to transitions \emph{within} a $K$-th order model, we obtain a multi-order model that includes transitions between the nodes of a $k$-th order model and the nodes of a $k+1$-th order model for all orders $k<K$.
Transitions from/to initial state $*$ and terminal state $\dagger$ explicitly model the start and end points of paths, which is the basis for an end-to-end, out-of-sample prediction of full paths.

\begin{figure*}[htb!]
	\input{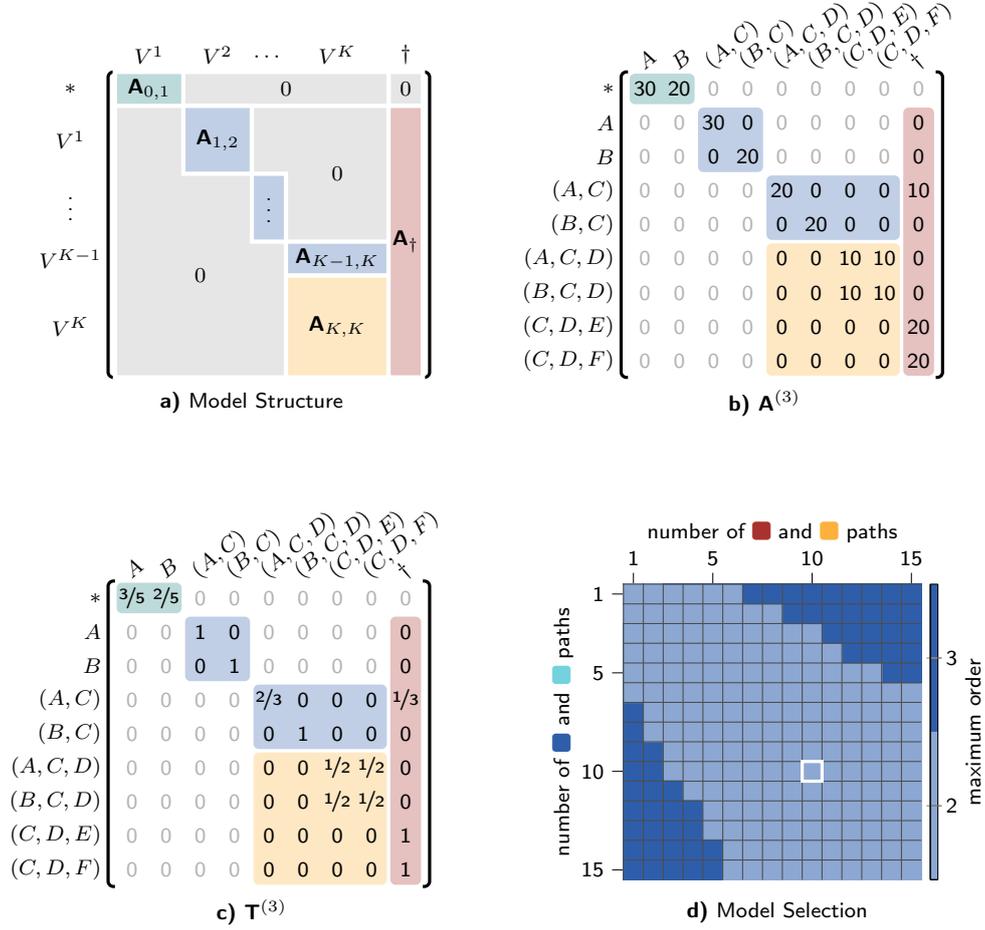}
	\caption{Multi-order matrix representation of MOGen. (a) shows the block structure of the multi-order adjacency matrix. (b) and (c) show $\textbf{A}$ and $\textbf{T}$ for MOGen with $K=3$ on the example from \cref{fig:infographic}a, where all paths occur 10 times. (d) shows the $K$ detected by model selection for varying observation counts. The observation from (b) and (c) is highlighted in white.
}\label{fig:model representation}
\end{figure*}

We can mathematically represent such a multi-order network model with maximum order $K$ by means of a \emph{multi-order adjacency matrix} $\textbf{A}^{(K)}$, whose overall structure is shown in \cref{fig:model representation}a.
This representation is inspired by so-called \emph{supra-adjacency matrices}, which have recently been proposed to represent the interconnected topologies of multi-layer networks~\cite{Kivela_MultiLayer}.
A multi-order adjacency matrix consists of blocks $\mathbf{A}_{k-1,k}$, whose entries count the observed transitions between $k-1$-th and $k$-th order nodes in a given multi set $S$ of paths.
The entries in the top-left block $\textbf{A}_{0,1} \in \mathbf{R}^{1\times n}$ count transitions $(*, v)$ for all nodes $v \in V$, i.e., the distribution of start nodes for all paths. 
The entries of the bottom-right block $\mathbf{A}^\dagger$ count \emph{terminal transitions} $(v_1, \mydots, v_k) \rightarrow \dagger$ for all orders $1 \leq k \leq K$.
For paths with lengths $l_i>K$, the entries in the bottom matrix $\mathbf{A}_{K,K}$ count the transitions between the nodes in a $K$-th order network model.
An example for a multi-order adjacency matrix that corresponds to a toy example where each path shown in in \cref{fig:infographic}a is observed 10 times is shown in~\cref{fig:model representation}a.
We note that structural zeros in $\textbf{A}^{(K)}$ (greyed out in \cref{fig:model representation}b) represent transitions that are \emph{impossible} based on the underlying network topology, which are different from zero entries that represent possible but \emph{unobserved} transitions (see zeros in highlighted blocks in ~\cref{fig:model representation}b).
The only model parameter $K$ determines the maximum order of the multi-order model, which resembles the maximum memory length in a Markov chain.
We note that, under the assumption that the paths in a multi set $S=\{ t_1, \ldots, t_m\}$ with maximum path length $l_{max}$ are generated independently of each other, the multi-order adjacency matrix $\textbf{A}^{(K)}$ is a lossless representation of $S$ iff $K \geq l_{max}$.

To facilitate prediction tasks, we use the multi-order adjacency matrix to define a \emph{probabilistic generative model} of variable-length paths.
This model is fully described by a \emph{multi-order transition matrix $\textbf{T}^{(K)}$}, which we obtain by a row normalisation of $\textbf{A}^{(K)}$.
This yields a stochastic matrix $\textbf{T}^{(K)}$, whose entries give the transition probabilities in a probabilistic model for paths.
The transition matrix of a generative multi-order model with maximum order $K=3$ that corresponds to the toy example in \cref{fig:infographic}a is shown in \cref{fig:model representation}c.
We are generally interested in maximally parsimonious models with maximum order $K \ll l_{max}$ that still capture relevant patterns in the sequence of nodes traversed by paths.
Different from the lossless representation $\textbf{A}^{(K)}$ obtained for $K>l_{max}$, we can view such a model as a lossy compression or summarisation of paths, based on the sequential patterns in $S$.
In the following, we show how we can adapt information-theoretic model selection to choose the optimal maximum order $K$ of a multi-order model that (i) yields an optimal summarisation of paths, and (ii) can be used to predict paths.

\subsection{Model Selection}
\label{sec:method:modelselection}

To choose a multi-order model with an optimal maximum order $K$ for a given set of paths $S$, we must first evaluate how well a given model describes a data set. 
Taking a statistical inference perspective, we first define the probability of a single observed path $t=v_1 \rightarrow \dots \rightarrow v_l$ given a multi-order model described by $\mathbf{T}^{(K)}$, as
\begin{align*}
{P}(t\vert \textbf{T}^{(K)}) =~&\textbf{T}^{(K)}[*,v_1] \cdot
\prod_{k=1}^{K-1} \textbf{T}^{(K)}[({v_1, \mydots, v_k}), (v_1, \mydots, v_{k+1})] \cdot \\
&\prod_{i=K+1}^{l} \hspace{-1.3mm} \textbf{T}^{(K)}[(v_{i-K}, \mydots, v_{i-1}), (v_{i-K+1}, \mydots, v_{i})] \cdot\\
&\cdot\textbf{T}^{(K)}[(v_{l-K+1}, \mydots, v_l), \dagger]
%\label{eq:likelihoodPath}
\end{align*}
A likelihood function for a multi-order model given a multi set $S$ of $m$ independent paths can then be obtained by taking the product over all path probabilities, i.e., 
\begin{align}
\mathcal{L}(\textbf{T}^{(K)}\vert S) = \prod_{t \in S}{P}(t\vert \textbf{T}^{(K)})
\label{eq:likelihoodData}
\end{align}
We highlight that the specific definition of the multi-order model allows the comparison of likelihoods across different maximum orders $K$.
This is due to the fact that the number of observed transitions that enter the likelihood calculation in \cref{eq:likelihoodData} is independent of the maximum order $K$, which is not the case for a standard higher-order Markov chain model.
We further note that, due to the inclusion of the special transitions to/from start and terminal states, the resulting models are not nested, i.e., a multi-order model with maximum order $K$ is not contained in the parameter space of a more complex model with maximum order $K'>K$.
While this difference to previous works on multi-order models is crucial to apply the model to the path prediction problem, it rules out the application of the likelihood-ratio based model selection proposed in~\cite{scholtes2017network}.

We address this problem with an information-theoretic approach to model selection. 
We specifically use Akaike's Information Criterion (AIC) \cite{Akaike1973_AIC} to balance the explanatory power of a multi-order model for a data set $S$ with the complexity of the model, which is measured in terms of the degrees of freedoms in the parameter space.
The AIC value can be interpreted as an estimate for the expected relative Kullback-Leibner (KL) divergence between the fitted model and the unknown true process that generates the observed data.
This corresponds to the expected loss of information in bits if we were to replace the unknown process that generates the data by our model.
The best model among a set of candidate models is thus the one that minimises Akaike's Information Criterion.
For a multi-order model, we can calculate the AIC as
\begin{align}
	\text{AIC}(\textbf{T}^{(K)}) = 2d(\textbf{T}^{(K)}) - 2\ln\left\{\mathcal{L}(\textbf{T}^{(K)}\vert S)\right\}
	\label{eq:AIC}
\end{align}
where $d(\textbf{T}^{(K)})$ are the degrees of freedom of the model defined by $\textbf{T}^{(K)}$.
Different from existing works that have used the AIC to determine the optimal order of higher-order Markov chain models of categorical sequence data~\cite{Strelioff2017_MarkovChains,Singer2014_Memory}, we must account for (i) structural zeros in the transition matrix $T^{(K)}$ that result from the fact that we model paths restricted to a network, (ii) the multi-layer structure of our models that combine multiple orders $k$, and (iii) the additional model parameters used to model the transitions from/to the initial and terminal states.
This implies that the degrees of freedom of a multi-order model of paths depend on the topology of the underlying network, which we can encode in terms of a binary adjacency matrix $\mathbfcal{A}$.\footnote{If we assume that all possible transitions along the links of the underlying network have been observed at least once, we can directly derive $\mathbfcal{A}$ from the training data.}
The possible paths of exactly length $k$ in such a network are encoded by the non-zero entries in the $k$-th power of the binary adjacency matrix, which allows us to calculate the degrees of freedom of a model with maximum order $K$ as:
\begin{align}
d(\textbf{T}^{(K)}) = \sum_{k=1}^K \sum_{i,j}\left(\mathbfcal{A}^{k}\right)_{ij} + \vert V\vert - 1
\end{align}
Ignoring constant terms that do not depend on the maximum order $K$, using \cref{eq:AIC} we can estimate the optimal maximum order $\hat{K}$ for a data set $S$ as follows:
\begin{align}
	\hat{K} := \argmin_{K} \sum_{k=1}^K \sum_{i,j}\left(\mathbfcal{A}^{k}\right)_{ij} - \ln\left\{\mathcal{L}(\textbf{T}^{(K)}\vert S)\right\}.
	\label{eq:optimal}
\end{align}

To illustrate our method in the running toy example shown in \cref{fig:infographic}a, here we present results for a simple experiment, in which we vary the observed frequencies of the five (colour-coded) paths shown in~\cref{fig:infographic}a.
We apply the model selection to different observed frequencies of paths and report the estimated optimal parameter $\hat{K}$.
Due to the topology of the network shown in \cref{fig:infographic}a, paths that start in node $A$ or $B$ and continue to $C$ and $D$ either terminate in node $E$ or node $F$.
In order to model paths whose probability to terminate in $E$ or $F$ depends on whether they start in $A$ or $B$, a multi-order model with maximum order three is needed.
Such a dependency is introduced if the paths $A \rightarrow C \rightarrow D \rightarrow E$ and $B \rightarrow C \rightarrow D \rightarrow F$ occur more or less frequently than the paths $B \rightarrow C \rightarrow D \rightarrow E$ and $A \rightarrow C \rightarrow D \rightarrow F$.
\Cref{fig:model representation}d shows that in this case our model selection algorithm yields the correct optimal order of three.
If, however, the terminal node of a path is independent of the start node, a model with order two is sufficient in this toy example.
This is the case if the frequencies of all four paths are similar, in which case our model selection algorithm yields the correct optimal order of two (cf. \cref{fig:model representation}d).
This is illustrated by the identical values in $\textbf{T}^{(3)}_{3,3}$ in \cref{fig:model representation}c.
We note that those results merely serve to illustrate our method in the running toy example.
In the coming section, we validate our method through a cross-validation experiment in empirical data sets.
The results of this experiment show that the best prediction performance is obtained for the parameter $\hat{K}$ that is estimated using the proposed model selection algorithm.
This implies that, different from existing techniques, the parameter $K$ can be directly estimated from the data, which effectively turns our method into a parameter-free approach. 

\begin{table*}[tb!]
	\caption{Summary statistics for the six data sets used to validate MOGen.}\label{tab:dataset_statistics}

	\sffamily
	\footnotesize
	\begin{tabularx}{\linewidth}{X@{\hspace{-5mm}}rrrrrrrrr}
	\toprule
	{} & \multicolumn{2}{c}{paths} & \multicolumn{4}{c}{nodes on path} & \multicolumn{3}{c}{network topology} \\\cmidrule(lr){2-3}\cmidrule(lr){4-7}\cmidrule(lr){8-10}
	{} &  \multicolumn{1}{c}{total} &  \multicolumn{1}{c}{unique} &  \multicolumn{1}{c}{mean} & \multicolumn{1}{c}{median} & \multicolumn{1}{c}{min} & \multicolumn{1}{c}{max} & nodes & links &   density \\
	\midrule
	\textbf{BMS1}: online retailer clickstreams \cite{gazelle} \hspace{8mm}  &                59'601 &                 18'473 &     2.51 &1& 1 & 267  &            497 &          15'387 &  6.24\%\\
	\textbf{FIFA}: FIFA World Cup 98 server logs \cite{arlitt2000workload}      &                20'450 &                 20'053 &    36.24&31 & 9 & 100 &           2'990 &          73'530 &  0.82\%\\
	\textbf{MSNBC}: news website clickstreams \cite{cadez2000visualization}               &               31'790 &                30'247 &     13.33&11 & 9 & 100 &             17 &            270 &  93.43\%\\
	\textbf{WIKI}: Wikispeedia clickstreams \cite{West2012} &                76'193 &                 68'784 &     6.25& 5 & 1 & 435 &           4'179 &          69'800 &  0.40\%\\
	\cmidrule(lr){1-10}
	\textbf{AIR}: US flight itineraries \cite{FLdata}             &               286'810 &                 60'228 &     4.19&5 & 2 & 14 &            175 &           1'598 &  5.24\%\\
	\textbf{TUBE}: London Tube itineraries \cite{LTdata}          &              4'295'731 &                 32'313 &     7.86&7 & 2 & 36 &            276 &            663 &  0.87\%\\
	\bottomrule
	\end{tabularx}
\end{table*}

\section{Experimental Validation}
\label{sec:validation}

We experimentally validate \methodname/ in six empirical data sets containing (i) user clickstreams on the Web, and (ii) travel itineraries of passengers in a train and an airline network:
BMS1 contains 59'601 clickstreams of customers of the web retailer \emph{Gazelle.com}~\cite{gazelle}.
FIFA captures 20'450 clickstreams of users of the '98 FIFA World Cup website~\cite{arlitt2000workload}.
MSNBC contains 31'790 clickstreams of page categories on the MSNBC news website~\cite{cadez2000visualization}.
WIKI captures navigation paths of 76'193 users in the Wikipedia article graph, who were playing the game Wikispeedia~\cite{West2012}.
AIR contains 286'810 flight itineraries of passengers travelling on routes between US airports in 2001~\cite{FLdata}.
TUBE captures 4'295'731 itineraries of London Tube passengers~\cite{LTdata}.
The raw data for all data sets are freely available online.
Summary statistics are provided in \cref{tab:dataset_statistics}.
The data sets BMS1, FIFA and MSNBC were obtained from the repository published with \cite{SPMF}\footnote{see \url{https://www.philippe-fournier-viger.com/spmf/index.php?link=datasets.php}}.

In the following sections, we validate our method as follows: 
Addressing an important application in sequence modelling, we first apply \methodname/ to predict the next node on a path, comparing its performance to six state-of-the-art methods. 
Highlighting a major advantage of our method over existing techniques, we then show that we can use the proposed model selection algorithm to compute the parameter $K$ for which our model offers the best performance directly from the data. 
This avoids computationally expensive parameter search algorithms that have been used by competing sequence prediction methods.
Showcasing another contribution of our work, we further evaluate our method in a scenario where we target an out-of-sample prediction of full paths.
We finally report results that show the scalability of our method in real data.

\subsection{Next-Element Prediction}
\label{sec:validation:prediction}

We first evaluate the performance of our method in a next-element prediction setting.
Given a prefix of $k$ nodes $(v_1, \mydots, v_k)$ previously traversed by a path, we want to predict which node $v_{k+1}$ in the network is visited next.
We evaluate our method in a cross-validation experiment for the six empirical data sets outlined above.
Leveraging the fact that our model captures terminal nodes on paths, we consider the special state $\dagger$ as a prediction of a path termination.
We include this in the evaluation, because the prediction of terminal nodes is crucial in data on variable-length path data, e.g., to predict where users exit a website, where passengers end their itinerary, or whether a purchase is made in an online shop.
We compare the performance of \methodname/ to state-of-the-art sequence prediction algorithms that have addressed next-element prediction in categorical sequences.
Our choice of baseline methods is informed by the review of \cite{next_element_review}, which compares the performance of state-of-the-art methods in several data sets.
The authors find that AKOM \cite{AKOM} outperforms other algorithms in three out of four tested data sets.
Conversely, \cite{CPT+} show that CPT+ outperforms AKOM in terms of the accuracy of the next element prediction.
We thus use both CPT+ and AKOM as baseline, utilising the implementations provided in \cite{SPMF}. 
To enable a fair comparison for empty prefixes, we modified the implementation of AKOM to return a prediction based on the frequency of elements in the training data. 
This is described in \cite{AKOM}, however the implementation in \cite{SPMF} returns an empty prediction instead.
CPT+ predicts the next element in a sequence based on an internal assignment of weights to candidate elements, i.e., the prediction is not based on probabilities.
We thus normalise the weights assigned by CPT+ to obtain a probabilistic prediction.
While this allows us to compare the performance of CPT+ with the other (probabilistic) methods, we highlight that the evaluation in \cite{CPT+} is based on prediction accuracy.
In addition to those sequence modelling techniques, we compare our method to a probabilistic prediction derived from (i) a random walk in a higher-order network with a single order $k$ (NET), and (ii) the multi-order graphical model (MOG) introduced in \cite{scholtes2017network}.
For $k=1$, NET is identical to a prediction generated by a random walk in the underlying network.
Note that, despite a similarity in name, the method proposed in the present work considerable differs from MOM.
First, MOM does not model the endpoint of paths, which limits its ability to predict variable-length paths.
Second, it uses a model fitting approach that estimates transition probabilities based on sub-path frequencies rather than the transition probabilities used in our approach. 
Finally, it uses a model selection that builds on a nested model structure, which does not hold for our modelling framework.
In addition to those network-based models, we use a naive (parameter-free) random baseline (RND) that simply predicts the next node based on the relative frequency of nodes in the training set, i.e., a prediction that neither uses a Markov chain nor the network topology.
For AKOM, MOM, NET, and MOGen we report the results for the optimal (maximum) order parameter obtained via a grid search.
For CPT+, we use the default parameters chosen in \cite{CPT+}.

We split the paths in each data set into a training (90\%) and validation (10\%) set and evaluate the prediction performance of all models in terms of the \emph{cross-entropy loss function}
\begin{align}
H(p,q) := - \sum_{v \in V} p(v) \log q(v)
\end{align}
where, for a given prefix $(v_1, \mydots, v_k)$, $q(v)$ denotes the probability of the next element (i.e., target node) $v$ obtained from a model trained on the training set and $p(v)$ is the true distribution underlying the data.
Since the true distribution is unknown for the empirical data set, we compute $p(v)$ from the distribution of actual next nodes in the validation set. 
This yields the log-loss function commonly used in the cross-validation of probabilistic multi-class prediction~\cite{bishop2006pattern}. 
In the first evaluation, we further consider predictions for multiple prefix lengths up to a maximum length of six.
We consider those as multiple samples and assign equal sample weights such that sample weights for any given target node with different prefix lengths sum to one.
For the resulting loss function, a value of zero corresponds to a perfect prediction, which would allow us to replace the true distribution in the data by the prediction without information loss.
Non-zero values quantify the information loss in bits, i.e., larger values correspond to worse prediction performance.

\begin{table}[bt!]\sffamily\footnotesize
	\caption{Next-element prediction performance of all models.}\label{tab:empirical_results}
	\begin{tabularx}{\linewidth}{l@{\hspace{6mm}}R@{\hspace{6mm}}R@{\hspace{6mm}}R@{\hspace{6mm}}R@{\hspace{6mm}}R@{\hspace{6mm}}R@{\hspace{6mm}}}
		\multicolumn{7}{l}{\textbf{a)} Cross-entropy loss [bit] for prefixes up to length 6} \\
		\toprule
		{} & \multicolumn{1}{c}{BMS1\hspace*{-5mm}} & \multicolumn{1}{c}{FIFA\hspace*{-5mm}} & \multicolumn{1}{c}{MSNBC\hspace*{-5mm}} & \multicolumn{1}{c}{WIKI\hspace*{-5mm}} & \multicolumn{1}{c}{AIR\hspace*{-5mm}} & \multicolumn{1}{c}{TUBE\hspace*{-4mm}} \\
		\midrule
		\textbf{MOGen} &   \textbf{6.22} &   \textbf{7.49} &   \textbf{2.86} &   \textbf{7.83} &   \textbf{4.20} &   \textbf{1.81} \\
        AKOM  &  19.51 &   8.67 &  6.07 &  14.37 &  13.27 &   7.09 \\
        CPT+  &  19.19 &  13.35 &  7.39 &  17.44 &  14.83 &  10.42 \\
		\cmidrule(lr){1-7}
		NET   &  19.51 &   8.84 &  6.19 &  14.36 &  13.34 &   7.58 \\
        MOM   &  19.51 &   8.66 &  6.07 &  14.36 &  13.20 &   6.35 \\
		\cmidrule(lr){1-7}
		%RNode &  20.17 &  12.00 &  12.72 &  17.09 &  15.59 &  12.82 \\
        RND &  19.56 &   9.69 &  6.78 &  15.81 &  14.12 &  11.94 \\
		\bottomrule \\[-2mm]
		\multicolumn{7}{l}{\textbf{b)} MOGen: detected and best performing maximum order} \\
		\toprule
		detected & \multicolumn{1}{c}{1\hspace*{-5mm}} & \multicolumn{1}{c}{1\hspace*{-5mm}} & \multicolumn{1}{c}{2\hspace*{-5mm}} & \multicolumn{1}{c}{1\hspace*{-5mm}} & \multicolumn{1}{c}{2\hspace*{-5mm}} & \multicolumn{1}{c}{6\hspace*{-5mm}} \\
		best     & \multicolumn{1}{c}{1\hspace*{-5mm}} & \multicolumn{1}{c}{1\hspace*{-5mm}} & \multicolumn{1}{c}{2\hspace*{-5mm}} & \multicolumn{1}{c}{1\hspace*{-5mm}} & \multicolumn{1}{c}{2\hspace*{-5mm}} & \multicolumn{1}{c}{6\hspace*{-5mm}} \\
		\bottomrule
	\end{tabularx}
\end{table}

%{} & \rotatebox{30}{BMS1} & \rotatebox{30}{FIFA} & \multicolumn{1}{c}{MSNBC} & \multicolumn{1}{c}{WIKI} & \multicolumn{1}{c}{AIR} & \multicolumn{1}{c}{TUBE} \\

In \cref{tab:empirical_results}a we report the cross-entropy loss (in bits) for the six empirical data sets and the six prediction methods described above.
For each algorithm, we report the performance for the parameters that yield the smallest cross-entropy loss.
The results show that MOGen outperforms all other methods for all of the six data sets.
For most of the data sets, we further observe a considerable difference in the cross-entropy loss function.
We hypothesise that this large difference in performance is due to the fact that different from other methods MOGen (i) explicitly models varying-length paths in a network, and (ii) is able to predict the termination of paths.
This highlights the main contribution of our work, which is a model that specifically accounts for the characteristics of data on paths in networks.
We further note that MON and NET yield identical results for BMS1 and WIKI.
The reason for this is that in those two data sets both MOM and NET yield the best prediction performance for a (maximum) order of one, in which case a multi-order model is identical to a first-order model.
We observe the smallest cross-entropy loss for MSNBC, AIR, TUBE, which we hypothesise is due to the ---relative to the size of the underlying network--- large number of observed paths.
FIFA, WIKI, and BMS1 generally yield the largest cross-entropy loss across all methods, which is likely to be  due to the large network topologies and a relatively small number of observations, which hinders a reliable detection of generalisable sequential patterns.

We finally observe that CPT+ shows the worst performance of all methods and it is even outperformed by the random baseline in all but one data sets.
This is likely due to the fact that we evaluate the methods based on probabilities assigned to the next elements, which is not the originally intended use of CPT+.
This is in line with the fact that a recent review found that the performance of CPT+ is inferior to the probabilistic method AKOM, while \cite{CPT+} have found the \emph{accuracy} of predictions generated by CPT+ to be superior to AKOM.

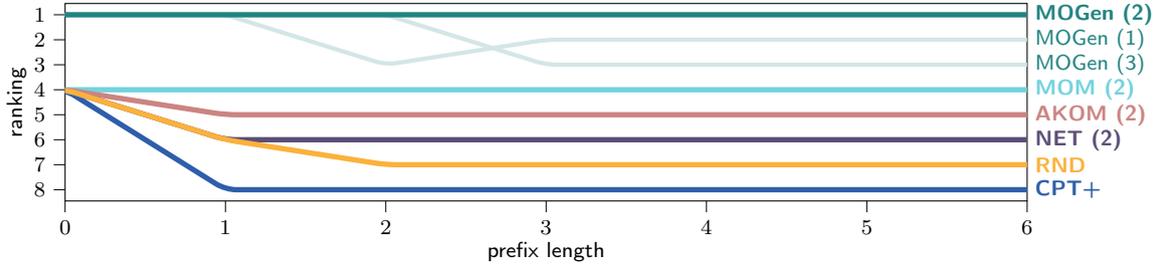
\begin{figure}[bt!]
	\centering
	\begin{tikzpicture}\sffamily\footnotesize
	
	\definecolor{color0}{rgb}{0.129411764705882,0.513725490196078,0.501960784313725}
	\definecolor{color1}{rgb}{0.356862745098039,0.305882352941176,0.466666666666667}
	\definecolor{color2}{rgb}{0.450980392156863,0.823529411764706,0.870588235294118}
	\definecolor{color3}{rgb}{0.180392156862745,0.368627450980392,0.666666666666667}
	\definecolor{color4}{rgb}{0.658823529411765,0.196078431372549,0.176470588235294}
	\definecolor{color5}{rgb}{0.984313725490196,0.694117647058824,0.235294117647059}
	
	\begin{axis}[
	width=.89\linewidth,
	height=4.2cm,
	tick align=outside,
	tick pos=left,
	x grid style={white!69.01960784313725!black},
	xlabel={prefix length},
	xmin=0, xmax=6,
	xtick style={color=black},
	y dir=reverse,
	y grid style={white!69.01960784313725!black},
	ylabel={ranking},
	ymin=0.55, ymax=8.45,
	ytick style={color=black},
	title style={yshift=-2mm},
	ylabel style={yshift=-1mm},
	xlabel style={yshift=1mm},
	clip mode=individual,
	xtick={0,1,2,3,4,5,6},
	ytick={1,2,3,4,5,6,7,8,9,10},
	]
	\addplot [rounded corners, ultra thick, color0!20, forget plot]
	table {%
		-1 1
		0 1
		1 1
		2 3
		3 2
		4 2
		5 2
		6 2
	} node[pos=1] (MOGen1) {};
	\addplot [rounded corners, ultra thick, color0!20, forget plot]
	table {%
		-1 1
		0 1
		1 1
		2 1
		3 3
		4 3
		5 3
		6 3
	} node[pos=1] (MOGen3) {};
	\addplot [rounded corners, line width=2pt, color0, opacity=1, forget plot]
	table {%
		-1 1
		0 1
		1 1
		2 1
		3 1
		4 1
		5 1
		6 1
	} node[pos=1] (MOGen2) {};
	\addplot [rounded corners, line width=2pt, color1, opacity=1, forget plot]
	table {%
		-1 4
		0 4
		1 6
		2 6
		3 6
		4 6
		5 6
		6 6
	} node[pos=1] (NET) {};
	\addplot [rounded corners, line width=2pt, color2, opacity=1, forget plot]
	table {%
		-1 4
		0 4
		1 4
		2 4
		3 4
		4 4
		5 4
		6 4
	} node[pos=1] (MOM) {};
	\addplot [rounded corners, line width=2pt, color3, opacity=1, forget plot]
	table {%
		-1 4
		0 4
		1 8
		2 8
		3 8
		4 8
		5 8
		6 8
	} node[pos=1] (CPT) {};
	\addplot [rounded corners, line width=2pt, color4!60, opacity=1, forget plot]
	table {%
		-1 4
		0 4
		1 5
		2 5
		3 5
		4 5
		5 5
		6 5
	} node[pos=1] (AKOM) {};
	\addplot [rounded corners, line width=2pt, color5, opacity=1, forget plot]
	table {%
		-1 4
		0 4
		1 6
		2 7
		3 7
		4 7
		5 7
		6 7
	} node[pos=1] (RND) {};
	
	\node[anchor=west] at (MOGen1) {\footnotesize \color{customG} MOGen (1)};
	\node[anchor=west] at (MOGen2) {\footnotesize \textbf{\color{customG} MOGen (2)}};
	\node[anchor=west] at (MOGen3) {\footnotesize \color{customG} MOGen (3)};

	\node[anchor=west] at (AKOM) {\footnotesize \textbf{\color{colorSG!60} AKOM (2)}};
	\node[anchor=west] at (CPT) {\footnotesize \textbf{\color{customB} CPT+}};
	\node[anchor=west] at (NET) {\footnotesize \textbf{\color{customP} NET (2)}};
	\node[anchor=west] at (MOM) {\footnotesize \textbf{\color{customT} MOM (2)}};
	\node[anchor=west] at (RND) {\footnotesize \textbf{\color{customY} RND}};
	
	\end{axis}
	
	%\node at (axis cs:6,1) {test};
	
	\end{tikzpicture}
	\caption{Prediction performance on different prefix lengths for MSNBC.}\label{fig:prefix_lengths}
\end{figure}

\Cref{tab:empirical_results} reports the prediction performance of methods across multiple prefix lengths and for the parametrisation that yields the best performance.
A further interesting question is how the performance of methods compares for different prefix lengths and different parameters.
To answer this question, in \cref{fig:prefix_lengths} we show a ranking of methods for different prefix lengths used in the prediction in the MSBNC data set.
For MOGen we show the performance of all considered parameters from $K=1$ to $K=3$ (MOGen(1) to MOgen(3)), while for other methods we show order parameter that yields the best performance (shown in brackets).
The results show that MOGen outperforms other methods across all prefix lengths and for all parameters $K$.

\subsection{Validation of Model Selection}
\label{sec:validation:modelselection}

All results reported in the previous section have been obtained by performing a grid-search of model parameters for which the models provide the best prediction performance.
In the case of MOGen, the only parameter of our method is the maximum order $K$ of the multi-order model.
As explained previously, our method provides a model selection algorithm that enables us to estimate the optimal value of the maximum order $\hat{K}$ directly from the data.
This avoids the need for a computationally expensive grid search for the optimal parameter and effectively makes our method parameter-free, which is a major advantage over competing methods.
In \cref{tab:empirical_results}b, we report the output of the AIC-based model selection of MOGen for the six empirical data set (top row), comparing it to the maximum order $K$ that yields the best prediction performance (bottom row).
For the clickstream data BMS1, FIFA, and WIKI we obtain an optimal maximum order of $K=1$. 
For MSNBC, AIR, and TUBE we obtain optimal values of $K=2$ and $K=6$ respectively, which highlights the need for a higher-order model to capture sequential patterns.
For all analysed data sets, the optimal order detected by the model selection provides the best prediction performance in a cross-validation, which confirms that our method yields models that neither underfit nor overfit sequential patterns in paths.

\subsection{Out-Of-Sample Prediction of Paths}
\label{sec:validation:forecasting}

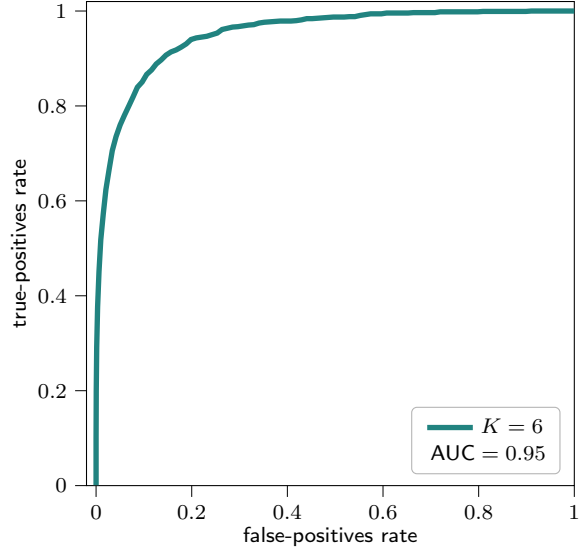
\begin{figure}
\centering
	% This file was created by tikzplotlib v0.9.1.
\begin{tikzpicture}\sffamily\footnotesize

\begin{axis}[
width=8cm,
height=8cm,
legend cell align={left},
legend style={fill=none, text opacity=1, at={(0.82,0.12)}, anchor=center, draw=none},
tick align=outside,
tick pos=left,
x grid style={white!69.0196078431373!black},
xlabel={false-positives rate},
xmin=-.02, xmax=1,
xtick style={color=black},
y grid style={white!69.0196078431373!black},
ylabel={true-positives rate},
ymin=0, ymax=1.02,
ytick style={color=black},
title style={yshift=-2mm},
xlabel style={yshift=1mm},
ylabel style={yshift=-1mm},
]
\addplot [line width=2pt, customG]
table {%
0 0.00084530853761623
0 0.101437024513948
0.000375586854460094 0.19949281487743
0.00131455399061033 0.291631445477599
0.0031924882629108 0.376162299239222
0.00610328638497653 0.450549450549451
0.00985915492957747 0.518174133558749
0.0150234741784038 0.572273879966188
0.0205633802816901 0.623837700760778
0.0271361502347418 0.665257819103973
0.0338967136150235 0.705832628909552
0.0417840375586854 0.73541842772612
0.0503286384976526 0.759932375316991
0.0592488262910798 0.78021978021978
0.0683568075117371 0.799661876584953
0.0773708920187793 0.819103972950127
0.0863849765258216 0.839391377852916
0.0963380281690141 0.850380388841927
0.105821596244131 0.866441251056636
0.115962441314554 0.875739644970414
0.125821596244131 0.888419273034658
0.136056338028169 0.89687235841082
0.146197183098592 0.907016060862215
0.156619718309859 0.913778529163145
0.167417840375587 0.918005071851226
0.177934272300469 0.923922231614539
0.188450704225352 0.930684699915469
0.198591549295775 0.939983093829248
0.209483568075117 0.943364327979713
0.220469483568075 0.945054945054945
0.231549295774648 0.946745562130177
0.242347417840376 0.950126796280642
0.253239436619718 0.953508030431107
0.263568075117371 0.961115807269653
0.274553990610329 0.963651732882502
0.285446009389671 0.966187658495351
0.296619718309859 0.967032967032967
0.307605633802817 0.968723584108199
0.31868544600939 0.970414201183432
0.329765258215962 0.971259509721048
0.340657276995305 0.974640743871513
0.351643192488263 0.976331360946746
0.362816901408451 0.977176669484362
0.373896713615023 0.978021978021978
0.385070422535211 0.978867286559594
0.396244131455399 0.978867286559594
0.407511737089202 0.978867286559594
0.418591549295775 0.979712595097211
0.429671361502347 0.981403212172443
0.44056338028169 0.983939137785292
0.451830985915493 0.983939137785292
0.462910798122066 0.984784446322908
0.474084507042254 0.985629754860524
0.485164319248826 0.98647506339814
0.496338028169014 0.987320371935757
0.507511737089202 0.987320371935757
0.518779342723005 0.987320371935757
0.529859154929577 0.988165680473373
0.54112676056338 0.988165680473373
0.552018779342723 0.990701606086221
0.563098591549296 0.992392223161454
0.574084507042254 0.994082840236686
0.585352112676056 0.994082840236686
0.596525821596244 0.994082840236686
0.607605633802817 0.995773457311919
0.618779342723005 0.995773457311919
0.630046948356808 0.995773457311919
0.641220657276995 0.995773457311919
0.652488262910798 0.995773457311919
0.663568075117371 0.996618765849535
0.674835680751174 0.996618765849535
0.686009389671361 0.996618765849535
0.697276995305164 0.996618765849535
0.708450704225352 0.996618765849535
0.719530516431925 0.998309382924768
0.730704225352113 0.998309382924768
0.741971830985915 0.998309382924768
0.753145539906103 0.998309382924768
0.764413145539906 0.998309382924768
0.775586854460094 0.998309382924768
0.786854460093897 0.998309382924768
0.798028169014084 0.998309382924768
0.809201877934272 0.999154691462384
0.82037558685446 0.999154691462384
0.831643192488263 0.999154691462384
0.842816901408451 0.999154691462384
0.854084507042253 0.999154691462384
0.865258215962441 0.999154691462384
0.876525821596244 0.999154691462384
0.887699530516432 0.999154691462384
0.898967136150235 0.999154691462384
0.910046948356808 1
0.92131455399061 1
0.932488262910798 1
0.943755868544601 1
0.954929577464789 1
0.966197183098592 1
0.977370892018779 1
0.988638497652582 1
0.999906103286385 1
};
\addlegendentry{$K=6$}

\node at (axis cs: .82,.07) {$\text{AUC}=0.95$};
\draw[black!30, rounded corners=2] (axis cs: .66,.165) rectangle (axis cs:.97,.025);

\end{axis}

\end{tikzpicture}
	\caption{Out-of-sample path prediction with MOGen. ROC and AUC for a 99/1 training/validation split for TUBE.}\label{fig:path_prediction}
\vspace{-.5cm}
\end{figure}

An important motivation of our method is its ability to perform an out-of-sample prediction of full paths based on a training set.
Different from the next-element prediction addressed earlier, this addresses applications where we have access to a set of paths that allows us to learn a multi-order generative model.
The task is then to predict a set of variable-length paths (from the start to the terminal nodes) that occur in a validation set.
This problem has a number of interesting applications, e.g., speculative caching in web servers, the prediction of information cascades, or demand prediction in transportation systems based on past passenger itineraries.
Addressing the latter scenario, we use \methodname/ to predict passenger itineraries in the London Tube based on a training set.
We evaluate the performance of our method in terms of a binary classification experiment, where we use a training set to learn an optimal multi-order model and then use the learned model to predict the most frequently occurring paths in a validation set.
The prediction of most frequent paths utilises the fact that our method learns a generative model that includes the start and end points of paths, i.e., we use our model to generate a new set $S'$ of paths with variable lengths.
We rank paths in the generated set $S'$ based on their frequency and use the top $N$ paths as prediction of the top $10 \%$ most frequent paths in the validation set.
We interpret this prediction as a binary classification and repeat the prediction for different discrimination thresholds $N$.
We then compute true positive/false positive rates for different values of $N$, which can be used to compute the area under a ROC (receiver operating characteristic) curve.
The result of this out-of-sample prediction is shown in \cref{fig:path_prediction}.
The ROC curve indicates a high accuracy of our method, which is substantiated by a resulting AUC score of $0.95$.

\subsection{Scalability}
\label{sec:validation:scalability}
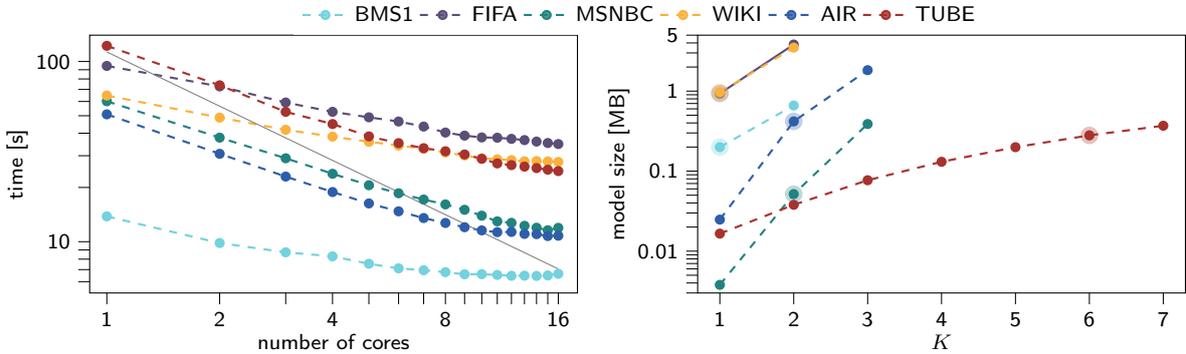
\begin{figure}
\centering
	% This file was created by tikzplotlib v0.8.7.

\centering
\begin{tikzpicture}\sffamily\footnotesize

\definecolor{color0}{rgb}{0.658823529411765,0.196078431372549,0.176470588235294}
\definecolor{color1}{rgb}{0.356862745098039,0.305882352941176,0.466666666666667}
\definecolor{color2}{rgb}{0.129411764705882,0.513725490196078,0.501960784313725}
\definecolor{color3}{rgb}{0.984313725490196,0.694117647058824,0.235294117647059}
\definecolor{color4}{rgb}{0.180392156862745,0.368627450980392,0.666666666666667}
\definecolor{color5}{rgb}{0.450980392156863,0.823529411764706,0.870588235294118}

\begin{axis}[
width=.5\linewidth,
height=5cm,
at={(0,0)},
legend cell align={left},
legend columns=6,
legend style={fill opacity=0.8, draw opacity=1, text opacity=1, at={(1.13,1)}, anchor=south, draw=none, font=\footnotesize},
log basis x={10},
log basis y={10},
tick align=outside,
tick pos=left,
x grid style={white!69.01960784313725!black},
xlabel={number of cores},
xmin=.9, xmax=18,
xmode=log,
xtick style={color=black},
y grid style={white!69.01960784313725!black},
ylabel={time [s]},
ymin=5.19253757988356, ymax=140,
ymode=log,
ytick style={color=black},
ylabel style={yshift=-1mm},
ytick={5,6,7,8,9,10,20,30,40,50,60,70,80,90,100,200,300,400,500,600,700,800},
yticklabels={,,,,,10,,,,,,,,,100,,,,,,,800},
xtick={1,2,3,4,5,6,7,8,9,10,11,12,13,14,15,16},
xticklabels={1,2,,4,,,,8,,,,,,,,16},
xlabel style={yshift=1mm},
]
\addplot [thick, color5, dashed, mark=*, mark size=1.5, mark options={solid}]
table {%
	1 13.8192643165588
	2 9.83069424629211
	3 8.73187561035156
	4 8.29330639839172
	5 7.54267654418945
	6 7.11409673690796
	7 6.94424166679382
	8 6.77461047172546
	9 6.59391198158264
	10 6.60235991477966
	11 6.53785667419434
	12 6.46859474182129
	13 6.47811193466187
	14 6.4557785987854
	15 6.49782958030701
	16 6.63678874969482
};
\addlegendentry{BMS1}
\addplot [thick, color1, dashed, mark=*, mark size=1.5, mark options={solid}]
table {%
	1 94.508536195755
	2 72.7419847488403
	3 59.206876707077
	4 52.565966463089
	5 48.9997931957245
	6 46.4564309120178
	7 43.5418328762054
	8 40.3387373924255
	9 38.7869194984436
	10 37.8975628852844
	11 37.7611772537231
	12 37.2216829299927
	13 36.6264272689819
	14 35.9489428520203
	15 35.3023224830627
	16 34.8326632499695
};
\addlegendentry{FIFA}
\addplot [thick, color2, dashed, mark=*, mark size=1.5, mark options={solid}]
table {%
	1 60.1577798366547
	2 37.7974452018738
	3 29.035911655426
	4 23.8229816913605
	5 20.5762917041779
	6 18.5710159778595
	7 17.1389587879181
	8 16.102964925766
	9 15.0330695152283
	10 13.9483651161194
	11 12.9711452960968
	12 12.7250426292419
	13 12.2505511283875
	14 11.9523522377014
	15 11.5730309009552
	16 11.9374317646027
};
\addlegendentry{MSNBC}
\addplot [thick, color3, dashed, mark=*, mark size=1.5, mark options={solid}]
table {%
	1 64.7531494617462
	2 48.8164441108704
	3 41.7781118869782
	4 38.2797618865967
	5 35.8415305137634
	6 34.1666696071625
	7 32.7692665100098
	8 31.2634082317352
	9 30.0997522354126
	10 29.0207458972931
	11 28.6477996826172
	12 28.3479271888733
	13 27.928352022171
	14 28.0018908500671
	15 27.8793536186218
	16 27.7191650867462
};
\addlegendentry{WIKI}
\addplot [thick, color4, dashed, mark=*, mark size=1.5, mark options={solid}]
table {%
	1 50.9003230571747
	2 30.7465590953827
	3 22.9819120883942
	4 18.8631848812103
	5 16.2997712135315
	6 14.7420334339142
	7 13.5269356250763
	8 12.7033568382263
	9 12.0312973499298
	10 11.5531391620636
	11 11.2958755970001
	12 11.3298570632935
	13 11.0479576587677
	14 11.0098202228546
	15 10.7480888843536
	16 10.796962594986
};
\addlegendentry{AIR}
\addplot [thick, color0 , dashed, mark=*, mark size=1.5, mark options={solid}]
table {%
	1 122.165865182877
	2 73.7851763248444
	3 52.5602885246277
	4 44.9718235492706
	5 38.4011479854584
	6 35.2319777011871
	7 33.0370772361755
	8 31.7685067176819
	9 30.4977961063385
	10 28.8314204692841
	11 27.1905256271362
	12 26.6060252666473
	13 26.1021313667297
	14 25.6239795684814
	15 25.0918793201447
	16 24.654498052597
};
\addlegendentry{TUBE}
\addplot [black!50, forget plot]
table {%
	1 113
	2 56.5
	3 37.6666666666667
	4 28.25
	5 22.6
	6 18.8333333333333
	7 16.1428571428571
	8 14.125
	9 12.5555555555556
	10 11.3
	11 10.2727272727273
	12 9.41666666666667
	13 8.69230769230769
	14 8.07142857142857
	15 7.53333333333333
	16 7.0625
};
\end{axis}

\begin{axis}[
width=.5\linewidth,
height=5cm,
at={(.5\linewidth,0)},
%legend cell align={left},
%legend columns=6,
%legend style={fill opacity=0.8, draw opacity=1, text opacity=1, at={(0.5,1.1)}, anchor=north west, draw=none},
log basis y={10},
tick align=outside,
tick pos=left,
x grid style={white!69.01960784313725!black},
xlabel={$K$},
xmin=0.7, xmax=7.3,
xtick style={color=black},
y grid style={white!69.01960784313725!black},
ylabel={model size [MB]},
ymin=0.003, ymax=5,
ymode=log,
ytick style={color=black},
ylabel style={yshift=-1mm},
ytick={0.001,0.002,0.003,0.004,0.005,0.006,0.007,0.008,0.009,0.01,0.02,0.03,0.04,0.05,0.06,0.07,0.08,0.09,0.1,0.2,0.3,0.4,0.5,0.6,0.7,0.8,0.9,1,2,3,4,5},
yticklabels={0.001,,,,,,,,,0.01,,,,,,,,,0.1,,,,,,,,,1,,,,5},
xtick={1,2,3,4,5,6,7},
xticklabels={1,2,3,4,5,6,7},
xlabel style={yshift=1mm},
]
\addplot [semithick, color5, opacity=0.3, mark=*, mark size=3, mark options={solid}, forget plot]
table {%
	1 0.199492
};
\addplot [thick, color5, dashed, mark=*, mark size=1.5, mark options={solid}, forget plot]
table {%
	1 0.199492
	2 0.660876
};
\addplot [semithick, color1, opacity=0.3, mark=*, mark size=3, mark options={solid}, forget plot]
table {%
	1 0.93288
};
\addplot [thick, color1, mark=*, mark size=1.5, mark options={solid}, forget plot]
table {%
	1 0.93288
	2 3.820076
};
\addplot [semithick, color2, opacity=0.3, mark=*, mark size=3, mark options={solid}, forget plot]
table {%
	2 0.05164
};
\addplot [thick, color2, dashed, mark=*, mark size=1.5, mark options={solid}, forget plot]
table {%
	1 0.003792
	2 0.05164
	3 0.388008
};
\addplot [semithick, color3, opacity=0.3, mark=*, mark size=3, mark options={solid}, forget plot]
table {%
	1 0.965184
};
\addplot [thick, color3, dashed, mark=*, mark size=1.5, mark options={solid}, forget plot]
table {%
	1 0.965184
	2 3.492352
};
\addplot [semithick, color4, opacity=0.3, mark=*, mark size=3, mark options={solid}, forget plot]
table {%
	2 0.418192
};
\addplot [thick, color4, dashed, mark=*, mark size=1.5, mark options={solid}, forget plot]
table {%
	1 0.024784
	2 0.418192
	3 1.827236
};
\addplot [semithick, color0, opacity=0.3, mark=*, mark size=3, mark options={solid}, forget plot]
table {%
	6 0.279752
};
\addplot [thick, color0, dashed, mark=*, mark size=1.5, mark options={solid}, forget plot]
table {%
	1 0.01656
	2 0.037936
	3 0.07692
	4 0.130796
	5 0.199092
	6 0.279752
	7 0.369308
};
\end{axis}

\end{tikzpicture}\hfill
	\caption{Mean time for training model selection, as well as model size for all data sets (5 iterations). The grey line in the left panel indicates a linear speedup. Highlighted models in the right panel are the ones picked through model selection.}\label{fig:scalability}
\end{figure}
In a final experiment, we evaluate the scalability of our method.
Our implementation parallelises the training of a multi-order model by splitting paths into multiple sets and simultaneously computing entries of the multi-order adjacency matrix $\textbf{A}^{(K)}$ and transition matrix $\textbf{T}^{(K)}$ for each of the sets, and finally aggregating those entries.
We speed up the computation of the model likelihood (cf. \cref{eq:likelihoodData}) by computing the probabilities of paths in parallel.
The left panel of \cref{fig:scalability} shows the time (in seconds) that is required to (i) train multi-order models for multiple parameters $K$, and (ii) select the optimal parameter $\hat{K}$ using the method described in \cref{eq:optimal} for different numbers of processing cores\footnote{Results were obtained using a 16-core Intel Core i9-7960X processor.}.
The resulting model sizes (in MegaBytes) for different orders $K$ are shown in the right panel of \cref{fig:scalability}.

The results show that the time required to learn an optimal model primarily depends on the number of unique paths as well as their length (cf.~\cref{tab:dataset_statistics}).
We also observe a near linear reduction of the training time as we increase the number of available processing cores (grey line refers to an optimal linear speedup).
The deviation from the linear reduction is likely to originate from the non-parallel computation of the matrix powers of $\mathbfcal{A}$, which is required to compute the degrees of freedom.
The size of the models generally increases super-linearly with the maximum order $K$.
The scaling of model size depends on the size and algebraic connectivity~\cite{fiedler1973algebraic} of the underlying topology, which determines the growth of the degrees of freedom with the parameter $K$.
This is illustrated by the TUBE data set, where---due to a sparsely connected network topology---the increase in model size is lowest as the order increases.
All models selected by \methodname/ are below 1 MB in size.

\section{Conclusion}
\label{sec:conclusion}

The growing availability of temporal data provides interesting opportunities for new data science techniques that address the modelling and prediction of paths in networks.
We have shown that such data has unique characteristics that question the application of state-of-the-art sequential pattern mining, sequence modelling, and prediction algorithms. 
Moreover, data on paths contain information on the topology of interactions in networked systems that cannot be extracted with standard network analysis and graph mining techniques.
Addressing this gap, we propose \methodname/, a modelling and prediction framework that is based on multi-order generative models of paths observed in a network.
Different from existing sequence mining and modelling frameworks, \methodname/ explicitly models the start- and endpoints of variable-length paths as well as patterns in node sequences that are restricted by an underlying network topology.
The importance of those features is demonstrated by the fact that \methodname/ outperforms state-of-the-art sequence prediction techniques in terms of next-element prediction in six empirical data sets.
Going beyond previous works on categorical sequence prediction, we further show that our method can be used for the out-of-sample prediction of full paths, e.g., variable-length paths from start to end, using passenger itineraries in a real transportation network.
This has interesting applications, for example, in terms of demand prediction for smart mobility scenarios, that we currently explore in a collaboration with a transportation provider.
A major advantage of \methodname/ compared to competing methods is that it provides an information-based model selection algorithm to determine the optimal order parameter $K$ directly from data. 
An experimental evaluation in empirical data sets proves that \methodname/ is able to automatically learn models that generalise best to unseen data, which effectively turns it into a parameter-free method.
Apart from those contributions to the foundation of sequential pattern mining in temporal network data, we provide an Open Source parallel implementation of our method\footnote{Integrated in \url{https://github.com/pathpy/pathpy}} and evaluate its scalability in real data sets.

\section*{Acknowledgements}
The authors acknowledge discussions with Yan Zhang on an early version of this work.
The author Ingo Scholtes acknowledges funding through the Swiss National Science Foundation grant 176938 as well as through the project bergisch.smart.mobility, funded by the German state of North Rhine-Westphalia.

\bibliography{references}

\end{document}